\documentclass[11pt,a4paper]{article}
\usepackage[hyperref]{emnlp2020}
\usepackage{times}
\usepackage{latexsym}

\usepackage{microtype}
\usepackage{caption}
\usepackage{subcaption}
\usepackage{graphicx}
\usepackage{tikz}
\usetikzlibrary{trees,calc}
\aclfinalcopy 

\newcommand*\samethanks[1][\value{footnote}]{\footnotemark[#1]}
\title{Fair Embedding Engine: A Library for Analyzing and Mitigating Gender Bias in Word Embeddings}
\author{Vaibhav Kumar\thanks{ \hspace{0.1cm}  Authors have contributed equally.} \quad  Tenzin Singhay Bhotia\samethanks \quad Vaibhav Kumar\samethanks \\
        Delhi Technological University\\
        Delhi, India\\
        \texttt{\{kumar.vaibhav1o1, tenzinbhotia0, vaibhavk992\}@gmail.com} }
 
\begin{document}
\maketitle
\begin{abstract} 
Non-contextual word embedding models have been shown to inherit human-like stereotypical biases of gender, race and religion from the training corpora. To counter this issue, a large body of research has emerged which aims to mitigate these biases while keeping the syntactic and semantic utility of embeddings intact. This paper describes Fair Embedding Engine (FEE), a library for analysing and mitigating gender bias in word embeddings. FEE combines various state of the art techniques for quantifying, visualising and mitigating gender bias in word embeddings under a standard abstraction. FEE will aid practitioners in fast track analysis of existing debiasing methods on their embedding models. Further, it will allow rapid prototyping of new methods by evaluating their performance on a suite of standard metrics. 
\end{abstract}


\section{Introduction}
Non-contextual word embedding models such as Word2Vec \cite{mikolov2013distributed,mikolov2013efficient}, GloVe \cite{pennington2014glove} and FastText \cite{bojanowski2017fastText}  have been established as the cornerstone of modern natural language processing (NLP) techniques. The ease of usage followed by performance improvements \cite{turian-etal-2010-word} have made word embeddings pervasive across various NLP tasks. However, as with most things, the gains come at a cost, word embeddings also pose the risk of introducing unwanted stereotypical biases in the downstream tasks. \citet{bolukbasi2016man} showed that a Word2Vec model trained on the Google news corpus, when evaluated for the analogy \textit{man:computer programmer :: woman:?} results to the answer \textit{homemaker}, reflecting the stereotypical biases towards woman. Further, \citet{winobias} showed that models operating on biased word embeddings can leverage stereotypical cues in downstream tasks like co-reference resolution as heuristics to make their final predictions.

Addressing the issues of unwanted biases in learned word representations, recent years have seen a surge in the development of word embedding debiasing procedures. The fundamental aim of a debiasing procedure is to mitigate stereotypical biases while introducing minimal semantic offset, hence maintaining the usability of embeddings. Based upon the mode of operation, the debiasing methods can be classified into two categories: First, post-processing methods, which operate upon pre-trained word vectors \cite{bolukbasi2016man,kaneko2019gender, yang2020causal}. Second, learning based methods, which involve re-training the word embedding models by either making changes to the training data or to the training objective. \cite{zhao2018learning,lu2018gender,bordia2019identifying}. Along with the development of debiasing procedures, numerous metrics to evaluate the efficacy of each debiasing procedure have also been proposed \cite{zhao2018learning, bolukbasi2016man, kumar2020nurse}. Although the domain has largely benefited from the contributions of different researchers, the domain still lacks open source software projects that unify such diverse but fundamentally similar methods in an organized and standard manner. Therefore, the domain has a high barrier for newcomers to overcome, and the domain experts may still need to put in extra effort of building and maintaining their own codebases.

\begin{figure*}
    \centering
    \includegraphics[width=1.0\textwidth]{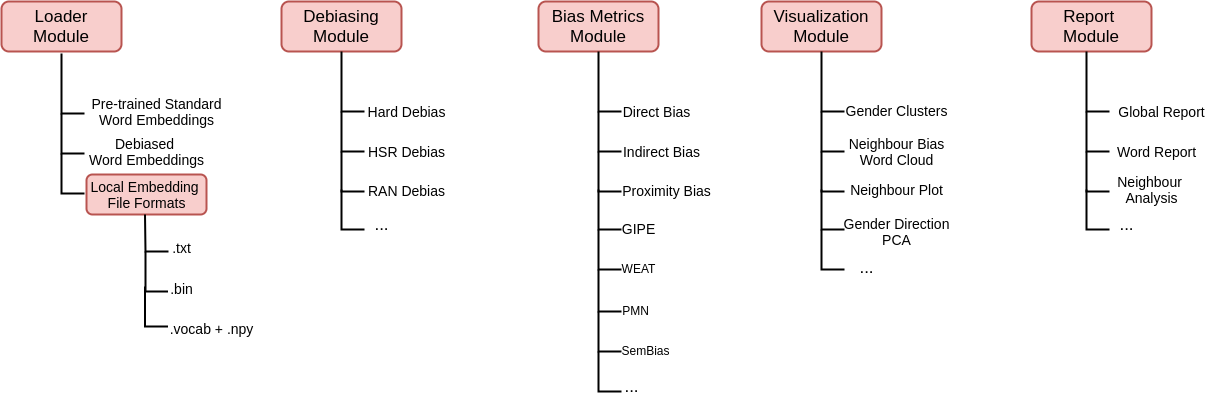}
    \caption{An inventory of implemented methods under four major modules that constitute FEE. Out of the box, FEE provides a subset of the prominent methods in each the modules. Further, each module can be easily extended to incorporate latest methods.}
    \label{fig:Fee_Structure}
\end{figure*}

To solve this problem, we introduce Fair Embedding Engine (FEE), a library which combines state of the art techniques for debiasing, quantifying and visualizing gender bias in non-contextual word embeddings for the English language. The goal of FEE is to serve as a unified framework towards the analysis of biases in word embeddings and the efficient development of better debiasing and bias evaluation methods. Conforming to the common style of implementation in some existing research works \cite{tolga, zhao2018learning}, we use Numpy ~\cite{oliphant2006guide, van2011numpy} arrays to store word vectors while keeping an index mapping to the strings of corresponding words. Further, we use the PyTorch ~\cite{paszke2017automatic} Autograd engine for gradient based optimization and Matplotlib ~\cite{Hunter:2007} for generating plots.
  FEE is made available at: \url{https://github.com/FEE-Fair-Embedding-Engine/FEE}.

\section{Related Work}
Since the study of stereotypical biases in NLP has received attention only in the recent years, the domain has not been a part of open source efforts that attempt to integrate diverse sets of independent methods. The 
only 
relevant 
open
source
software 
(OSS)
that 
we 
came 
across 
during 
our 
investigation 
was 
Word 
Embedding 
Fairness 
Evaluation 
(WEFE) 
framework \cite{badillawefe}.
For
a 
given 
collection
of
pre-trained 
word embedding
and
a
set 
of 
fairness
criteria,
WEFE
ranks 
the 
embeddings 
based 
on 
their 
performance 
on
an 
encapsulation 
of 
the 
fairness 
metrics. 
In order to achieve this ranking over an otherwise
disparate
set
of
fairness
metrics
(WEAT \cite{caliskan2017weat}, 
RND \cite{garg2018word}, 
and 
RNSB \cite{sweeney-najafian-2019-transparent})
WEFE
introduces
an
abstraction
which
generalizes
over
the
metrics
using 
a 
set 
of 
\textit{target} 
(the intended social class for which fairness is to be evaluated) 
and 
\textit{attribute} words
(the traits over which bias might exist for the selected target words).
Further,
\cite{badillawefe}
conclude
that
while 
existing 
fairness 
metrics 
show 
a 
strong 
correlation 
when 
used
for
evaluating
gender 
bias, 
only 
a 
weak 
correlation 
results 
when 
evaluating 
biases 
like 
religion 
and 
race.

Therefore, the focus of WEFE is limited to the evaluation of pre-trained word vectors on a suite of fairness metrics, lacking any support for debiasing methods. Further, only those evaluation metrics can be used which comply with the abstraction. FEE, on the other hand, provides holistic functionality by equipping a suite of evaluation and debiasing methods, along with a flexible design to assist researchers in developing new solutions. 

FEE currently offers three debiasing methods as a part of its debiasing module: HardDebias \cite{tolga}, HSRDebias \cite{yang2020causal}, and RANDebias \cite{kumar2020nurse}. The bias metrics module consist of the following: SemBias \cite{zhao2018learning}, direct and indirect bias \cite{bolukbasi2016man}, Gender-basied Illicit Proximity Estimate (GIPE) and Proximity bias \cite{kumar2020nurse}, Percent Male Neighbours (PMN) \cite{gonen2019lipstick} and Word Embedding Association Test (WEAT) \cite{caliskan2017semantics}. 

\section{Fair Embedding Engine}
 The core functionality of FEE is governed by five modules, namely \textit{Loader}, \textit{Debias}, \textit{Bias Metrics}, \textit{Visualization}, and \textit{Report}. Figure \ref{fig:Fee_Structure} illustrates the components for each module of FEE. In the following subsections, we delineate upon the implementation of each of the modules along with the motivation for their development. 

\subsection{Loader Module}
\textbf{Motivation}: The foremost step in the analysis of word embeddings is to load them into the random access memory. However, different formats of local embedding files, and heterogeneous formats of pre-trained embedding sources may entail disparate forms of access, making the loading process non-trivial. The loader module abstracts this pre-processing step and provides a standardized object based access of word embeddings to its users. 
\\
\textbf{Working}: The workhorse of the loader module is its Word Embedding class, \texttt{WE}. Any version of a word embedding model can be considered as a unique instance of the \texttt{WE} class. It consists of a user accessible \texttt{loader()} method that either takes in an embedding name representing a pre-trained word embedding, or a local embedding file path as input and returns an initialized \texttt{WE} object. We integrate the well established Gensim \cite{rehurek_lrec} API in our loader module for providing access to several pre-trained embeddings. However, since FEE focuses on the bias domain, it also provides the functionality to either store the debiased counterparts of Gensim-loaded embeddings, or load an externally downloaded debiased embedding file. For flexibility, the loader module supports three prominent file formats i.e. \texttt{.txt}, \texttt{.bin}, and \texttt{.vocab} (words) + \texttt{.npy} (vectors). Once, a \texttt{WE} object is initialized with an embedding version via the \texttt{loader()} method, a user can obtain the vector representation for a word by calling its vector method, \texttt{v()} with that word as its argument. All the subsequent modules of FEE operate on the \texttt{WE} object for achieving their objectives.  

\begin{figure*}[t]
\centering
\begin{subfigure}[b]{\textwidth}
   \centering
   \includegraphics[scale=0.43]{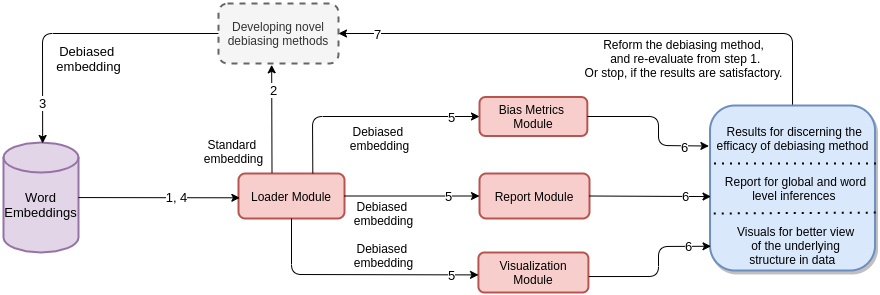}
   \caption{Developing novel debiasing methods}
   \label{fig:Ng1} 
\end{subfigure}
\begin{subfigure}[b]{\textwidth}
   \centering
   \includegraphics[scale=0.43]{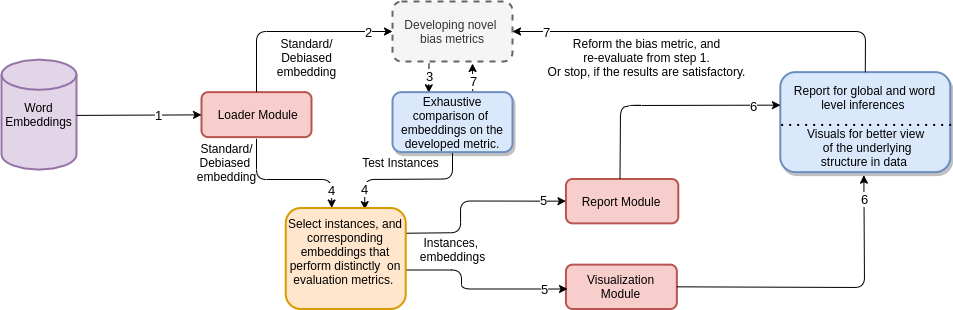}
   \caption{Developing novel bias metrics}
   \label{fig:Ng2} 
\end{subfigure}
\caption{FEE serves as a centralized resource for practitioners and researchers to develop novel debiasing methods and bias evaluation metrics. Figure (a) and (b) illustrate the possible workflow associated with each of the tasks respectively all made possible by the powerful abstraction provided by FEE.}
 \label{fig:development}
\end{figure*}

\subsection{Debiasing Module}
\textbf{Motivation}: The domain of bias in word representations considers effective debiasing methods as one of their ultimate objectives. Much effort has been made in the recent years to develop good debiasing methods. However, most works flaunt the efficacy of their debiasing procedures by applying them to a limited number of pre-trained embeddings. We hope that future works try to experiment their new methods or the existing ones on different embeddings. However, such a task involves refactoring and modification of individually tailored prior works. The debiasing module of FEE re-implements these diverse algorithms and provides a standardized access to users while facilitating reproducible research.\\
\textbf{Working}: The debiasing module of FEE currently provides access to some of the proposed post-processing debiasing procedures in the past, as shown in Figure \ref{fig:Fee_Structure}. Each debiasing method is represented by a unique class in the module. For instance, the Hard Debias methods proposed by \citet{bolukbasi2016man} is assigned a class named, \texttt{HardDebias}. Since all debiasing methods are fundamentally applied to a word embedding, the class of each debiasing method is initialised by a \texttt{WE} object. Each debiasing class has a common method called \texttt{run()} that takes in a list of words as argument and runs the entire debiasing procedure on it. As the debiasing procedure operates upon the \texttt{WE} object, the engineering effort in dealing with different embedding formats is mitigated. 

\subsection{Bias Metrics Module}
\textbf{Motivation}: Evaluation metrics provide the necessary quantitative support for comparing and contrasting between different debiasing methods. However, different research articles often show different results for the same metric despite having theoretically similar configurations. The bias metrics module of FEE is aimed at filling this gap, it provides a suite of bias metrics built on a common framework for facilitating reliable inference.\\
\textbf{Working}: The bias metrics module of FEE currently provides access to a number of evaluation metrics, as shown in Figure \ref{fig:Fee_Structure}. Each evaluation metric is represented by a unique class in the module. Each metric class depends on some common utilities and consists of multiple methods that implement their unique evaluation procedure. Similar to the debiasing module, each metric class's instance is initialised by a \texttt{WE} object. The metrics either operate on a single word, pair of words or list of words. Each metric class has a common method called \texttt{compute()} that returns the final result by accepting different arguments corresponding to the type of metric. The unified design of bias metric module fosters a standardized access to any bias based evaluation metric and facilitates reproducible research.

\subsection{Visualization Module}
\textbf{Motivation}: Visualizations provide useful insights into the behaviour of a set of data points. Many prior debiasing methods \cite{bolukbasi2016man, kumar2020nurse} have strongly motivated their work by illustrating certain undesirable associations prevalent in standard word embeddings. Thus, through \texttt{FEE} we also provide off the shelf visualization capabilities that might help users to build reliable intuitions and uncover hidden biases in their models.\\
\textbf{Working}: In this module, we implement a separate class for each visualization type. Just like other modules, a visualization class object is initialized by \texttt{WE} object, and makes use of some common utilities. Each visualization class has a \texttt{run()} method that takes in a word list and other optional arguments for producing the final visualizations. Figure \ref{fig:Fee_Structure} illustrates some off the shelf visualization options provided by FEE.

\subsection{Report Module}
\textbf{Motivation}: The bias metrics and visualization modules incorporate a plethora of components which provide an exhaustive set of results. However, sometimes a specific combination of their components can provide the needed information succinctly. Accordingly, the report module aims to provide a descriptive summary of bias in word embeddings at the word and global level.  \\
\textbf{Working}: The report module is comprised of two separate classes that are representative of a word and a global level report respectively. Both the classes operate on \texttt{WE} initialized embedding object and implement a common \texttt{generate()} method that creates a descriptive report. The \texttt{WordReport} class is useful for providing an abridged information about a single word vector in terms of bias. A call to the \texttt{generate()} method of \texttt{WordReport} utilizes the components of other modules and instantly reports the direct bias, proximity bias, neighbour analysis (\texttt{NeighboursAnalysis}), neighbour plot and a neighbour word cloud for a word.
The \texttt{GlobalReport} class, in contrast creates a concise report at the entire embedding level. Unlike the word level, \texttt{GlobalReport} class does not make use of the other modules, since it achieves all the required content from the embedding object. The \texttt{generate()} method of \texttt{WordReport} provides the information about $n$ most and least biased words in a word embedding space. 

\section{Developing new methods with FEE}
Despite the development of a large number of debiasing methods, the issue of bias in word representations still persists \cite{gonen2019lipstick} making it an active area of research. We believe that the design and wide variety of tools provided by FEE can play a significant role in assisting practitioners and researchers to develop better debiasing and evaluation methods. Figure \ref{fig:development} portrays FEE assisted workflows which abstract the routing engineering tasks and allow users to invest more time on the intellectually demanding questions. 

\section {Conclusion and future work}
In this paper, we described Fair Embedding Engine (FEE), a python library which provides central access to the state-of-the-art  techniques for quantifying, mitigating and visualizing gender bias in non-contextual word embedding models. We believe that FEE will facilitate the development and testing of debiasing methods for word embeddings. Further, it will make it easier to visualize the existing bias present in word vectors. In future, we would like to expand the capabilities of FEE towards contextual word vectors and also provide support towards biases other than gender and language other than English. We also look forward to integrate OSS such as WEFE \cite{badillawefe} to enhance the bias evaluation capabilities of FEE.
\bibliography{emnlp2020}
\bibliographystyle{acl_natbib}

\end{document}